\documentclass{article}

\usepackage[preprint]{neurips_2023}

\usepackage[utf8]{inputenc} 
\usepackage[T1]{fontenc}    
\usepackage{hyperref}       
\usepackage{url}            
\usepackage{booktabs}       
\usepackage{amsfonts}       
\usepackage{nicefrac}       
\usepackage{microtype}      
\usepackage{xcolor}         
\usepackage{graphicx}
\usepackage{amsmath}
\usepackage{amssymb}
\usepackage{algorithm}
\usepackage{algpseudocode}
\usepackage{subcaption}
\usepackage{lipsum}
\usepackage{wrapfig}

\title{A Probabilistic Approach to Self-Supervised Learning using Cyclical Stochastic Gradient MCMC}

\author{%
  Masoumeh Javanbakhat
    \\
    Digital Health-Machine Learning\\
  Hasso-Plattner-Institute, Germany \\
  \texttt{Masoumeh.Javanbakhat@hpi.de} \\
   \And
   Christoph Lippert \\
   Digital Health-Machine Learning\\
   Hasso-Plattner-Institute, Germany\\
   \texttt{Christoph.Lippert@hpi.de} \\
}

\begin{document}
\bibliographystyle{abbrvnat}
\maketitle

\begin{abstract}
In this paper we present a practical Bayesian self-supervised learning method with Cyclical Stochastic Gradient Hamiltonian Monte Carlo (cSGHMC). Within this framework, we place a prior over the parameters of a self-supervised learning model and use cSGHMC to approximate the high dimensional and multimodal posterior distribution over the embeddings. By exploring an expressive posterior over the embeddings, Bayesian self-supervised learning produces interpretable and diverse representations. Marginalizing over these representations yields a significant gain in performance, calibration and out-of-distribution detection on a variety of downstream classification tasks.
We provide experimental results on multiple classification tasks on four challenging datasets. Moreover, we demonstrate the effectiveness of the proposed method in out-of-distribution detection using the SVHN and CIFAR-10 datasets. 
\end{abstract}

\section{Introduction}\label{sec:intro}
Self-supervised learning is a learning strategy where the
data themselves provide the labels \citep{self-sup}. The aim of self-supervised learning is to learn useful representations of the input data without relying on human annotations \citep{Barlow}. Since they do not rely on annotated data, they have been used as an essential step in many areas such as natural language processing, computer vision and biomedicine \citep{handson}, where the data annotation is time-consuming and expensive.  
 
Despite the notable advancements made in recent years, self-supervised models are often trained using stochastic optimization methods which estimate the \textit{distribution} over parameters as a \textit{point mass}, ignoring the inherent uncertainty present in the parameter space. Remarkably, if the regularizer imposed on the model parameters is viewed as the the log of a prior on the distribution of the parameters, optimizing the cost function may be viewed as a \textit{maximum a-posteriori} (MAP) estimate of model parameters \citep{shape}. Bayesian methods provide principled alternatives that model the whole posterior over the parameters and effectively account for the inherent uncertainty in the parameter space \citep{cSGMCMC}. While the benefits of Bayesian methods and modeling uncertainty have been extensively explored in supervised learning \citep{precond,SWAG,Wilson_Izmailov}, their potential advantages in self-supervised learning remain largely unexplored. 

\begin{figure}[!t]
\begin{center}
\includegraphics[width=0.8\linewidth]{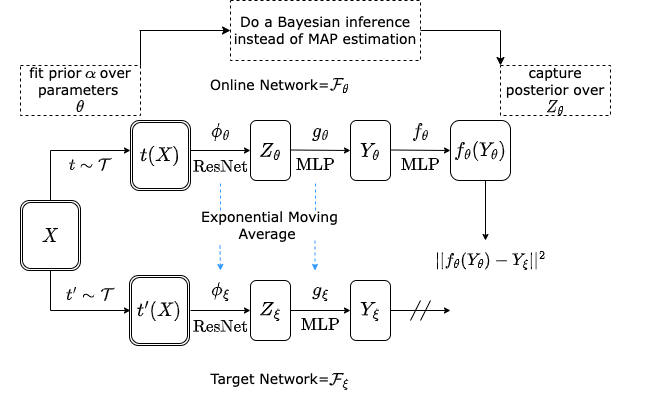}
\end{center}
\caption{An illustration of Bayesian BYOL. We fit a prior over parameters of encoder. Perform Bayesian optimization instead of MAP estimation, to capture posterior over embeddings. Marginalize over embeddings in downstream task.}
\label{pipeline}
\end{figure}

Indeed the posterior distribution over the parameters of a self-supervised learning model may be multimodal and thus insufficiently represented by a single point estimate. Each mode in the posterior can provide a meaningful different representation of data. By exploring the posterior distribution over the parameters instead of relying on point mass, our aim is to enhance performance and generalizability in downstream tasks. Additionally, it enables the estimation of uncertainties associated with predictions in downstream task, which holds significant value in numerous critical decision-making systems. 


 

\paragraph{Our contributions.}In this paper, we propose a novel Bayesian formulation for self-supervised learning that surpasses the limitations of MAP estimation by approximating the full posterior distribution over representations. To achieve this, we leverage the power of a family of Markov Chain Monte Carlo (MCMC) \citep{mcmc} methods known as Cyclical Stochastic Gradient Hamiltonian Monte Carlo (cSGHMC) \citep{cSGMCMC}, enabling us to effectively capture the multimodality inherent in the posterior distribution. Within this framework, we utilize BYOL \citep{BYOL}, a state-of-the-art model in contrastive learning, to learn representations. Our experimental results demonstrate the remarkable potential of Bayesian learning, which unlocks enhanced performance, superior generalizability, and improved calibration in various downstream tasks, including classification and out-of-distribution detection. Importantly, our approach also enables the estimation of uncertainty in the predictive space in downstream task, a crucial aspect that has been disregarded by deterministic nature of conventional self-supervised learning methods. 

\section{Related Works}\label{sec:related_work}
This work closely aligns with two lines of research: Bayesian inference and self-supervised learning.
\paragraph{Bayesian inference}
Bayesian Deep Learning evolving from Bayesian Neural Networks \citep{27,mcmc} provides a compelling alternative to point estimation by capturing model uncertainty or epistemic uncertainty. Sampling the posterior distribution poses challenges in general cases, leading to the adoption of approximation methods. Among these methods, MCMC algorithms stand out as a popular choice for accurately sampling the posterior distribution, while variational inference VI \citep{VI} offers a technique for learning an approximate posterior distribution. In recent research, Stochastic Gradient Markov Chain Monte Carlo (SG-MCMC) methods \citep{SGLD, SGHMC,crecip} have gained prominence for combining MCMC methods with minibatching, enabling scalable inference on large datasets. Additionally, a notable advancement in this domain is the introduction of Cyclical Stochastic Gradient MCMC (cSG-MCMC) \citep{cSGMCMC}. This method specifically addresses the exploration of highly multimodal parameter spaces within realistic computational budgets \citep{cSGMCMC}. 

\paragraph{Self Supervised Learning}
Self-supervised learning plays a crucial role in acquiring valuable representations from a vast amount of unlabeled data, leading to improved performance in downstream tasks. It is widely recognised as a pivotal step toward developing more capable and data-efficient learning systems \citep{Self_Iso}. Among the promising approaches in self-supervised learning, contrastive methods \citep{simCLR} stand out. These methods learn representations by maximizing the similarity between embeddings derived from different distorted versions of an image \citep{Barlow}. However, one major problem with similarity learning is feature collapse, where the learned features of the model converge to a single point in the feature space, resulting in a loss of discriminative power. To address this issue, several techniques have been proposed. For instance, in simCLR \citep{simCLR}, the use of negative samples is introduced. Another approach, employed in BYOL \citep{BYOL}, involves leveraging stop gradients to prevent feature collapse.

\paragraph{Pre-trianed models as Bayesian priors}
There have been previous works that formulate a pretrained representation as a Bayesian prior that is optimal for data from the second task. Notably \citet{deep_ref} extend the theory of reference priors to compute an uninformative Bayesian priors by maximizing the mutual information between the task and the weights. They apply reference priors in two problems: Bayesian semi-supervised learning using unlabeld data and transfer learning, where the labeled data from the source task are utilized. In another study, \citet{pretrain_loss} adopt a variational approach to construct an informative prior from pre-training data, aiming to maximize the performance in a single downstream task. While these approaches show promise, our method differs in several key aspects: (1) We use sampling to explore a full posterior over the representations, whereas \citet{pretrain_loss} rely on a variational approximation centered on one of the modes of the posterior. VI is prone to overlay representation even within the mode, potentially limiting its ability to capture uncertainty accurately. (2) our approach utilizes a simple yet effective representation for the posterior distribution, requiring minimal intervention while yielding promising results across various downstream tasks. It enables accurate and reliable uncertainty quantification which is crucial in many practical applications. In \citep{cSGMCMC} authors indicate the importance of capturing different modes in the posterior in order to accurately estimate uncertainties. By addressing these differences, our proposed approach offers a novel perspective on leveraging pretrained representations in a Bayesian framework, paving the way for improved performance and reliable uncertainty estimation in diverse downstream tasks. 


\section{Problem Statement}
Given a dataset $\mathcal{D}$, a self-supervised learning model $\mathcal{F}_{\theta}$ parameterized by $\theta$, aims to produce a representation $Z_{\theta}$ by solving a predefined proxy task.  
In this paper we wish to learn a distribution over the embeddings $Z_{\theta}$ by placing a prior over the parameters $\theta$ and adopting Bayesian learning instead of relying on MAP estimation. Our method is illustrated in Fig. \ref{pipeline}. To learn the representations, we use BYOL. In order to capture the distribution over the embeddings, we 
utilize cSGHMC. In the following sections, we first provide a description of the self-supervised learning model employed for representation learning. Then, we describe cSGHMC and highlight how it allows to obtain a distribution over the embeddings. 

\subsection{Self supervised learning}
\label{SSL}
The aim of contrastive learning is to learn representations by contrasting two augmented views of an image. Particularly BYOL learns representations by reducing a contrastive loss between two neural networks referred to as online network $\mathcal{F}_{\theta}$ (parameterized by $\theta$) and target network $\mathcal{F}_{\xi}$ (parameterized by $\xi$). Each network consists of three components, an encoder $\phi(.)$ (e.g., Resnet-18), a projection head $g(.)$ (e.g., an MLP) and a prediction head $f(.)$ (e.g., an MLP). For a given mini-batch $X=\{x_{i}\}_{i=1}^{N}$ sampled from a dataset $\mathcal{D}$ it produces two distorted views, $t(X)$ and $t^{\prime}(X)$, via a distribution of data augmentations $\mathcal{T}$. The two batches  of distorted views then are fed to the online network and the target network respectively, producing batches of embeddings, $Z_{\theta}$ and $Z_{\xi}$, respectively. These features are then transformed with the projection heads into $Y_{\theta}$ and $Y_{\xi}$. The online network then outputs a prediction $f_{\theta}(Y_{\theta})$ of $Y_{\xi}$ using prediction head $f_{\theta}$. Finally the following mean squared error between the normalized predictions $\bar{f}_{\theta}(Y_{\theta})$ and target projections $\bar{Y}_{\xi}$ is defined:

\begin{equation}
    \mathcal{L}_{\theta,\xi} = \| \bar{f_{\theta}}(Y_{\theta})-\bar{Y}_{\xi} \| ^{2} = 2- 2. \dfrac{\langle \bar{f_{\theta}}(Y_{\theta}), \bar{Y}_{\xi}\rangle}{\| \bar{f_{\theta}}(Y_{\theta})\|.\| \bar{Y}_{\xi}\|}. 
\end{equation}

$\tilde{\mathcal{L}}_{\theta,\xi}$ is computed by separately feeding $t^{\prime}(X)$ to the online network $\mathcal{F}_{\theta}$ and $t(X)$ to the target network $\mathcal{F}_{\xi}$. Indeed $\mathcal{L}_{\theta,\xi}$ and $\tilde{\mathcal{L}}_{\theta,\xi}$ are the same, only the views input to the target and online networks are swapped over. Then, at each training step, a stochastic optimization step is performed to minimize 
\begin{equation}
\label{lossB}
    \mathcal{L}^{\text{BYOL}}_{\theta, \xi} = \mathcal{L}_{\theta, \xi} + \tilde{\mathcal{L}}_{\theta,\xi}
\end{equation}

The gradient is taken only with respect to $\theta$. So, during training only the parameters $\theta$ are updated as follows:
\begin{equation}
\theta \leftarrow \text{optimizer}(\theta,\nabla_{\theta}\mathcal{L}_{\theta,\xi}^{\text{BYOL}}) 
\end{equation}
The weights $\xi$ are an exponential moving average of the online network's parameters $\theta$ with a target decay rate $\tau\in[0,1]$,
\begin{equation}
\xi  \leftarrow \tau \xi +(1-\tau) \theta .
\end{equation}
At the end of training, the encoder $\phi_{\theta}(.)$ is used for the downstream task. During training only the parameters $\theta$ of the online network $\mathcal{F}_{\theta}$ are updated. 

\subsection{Posterior Sampling using cSGHMC}
\label{cSGHMC}
In the Bayesian paradigm, for a given dataset $\mathcal{D}=\{x_{i}\}_{i=1}^{n}$ and a $\theta$-parameterized model, the following \textit{a-posterior distribution} over $\theta$ is computed using Bayes' rule as: $p(\theta| \mathcal{D})\propto p(\mathcal{D}|\theta)p(\theta)$, where $p(\theta)$ is a \textit{prior} assigned to the parameters $\theta$ and $p(\mathcal{D}|\theta)$ is the likelihood. 

In MAP optimization, the prior has the role of a regularizer and the likelihood has the role of a cost function. An optimizer is optimized to find the 
MAP solution which is amenable to the parameter update: 
\begin{equation}
\label{sgd}
   \Delta \theta = -\frac{\ell}{2}\left(\frac{n}{N}\sum_{i=1 }^{N}\nabla_{\theta} \log p(x_{i}|\theta)+ \nabla_{\theta}\log p(\theta)\right)
\end{equation}
for a given randomly sampled mini-batch $X=\{x_{i}\}_{i=1}^{N}\subset \mathcal{D}$ and learning rate $\ell$. 

In contrast to MAP optimization, in the Bayesian paradigm the model explores the distribution over the model parameters. \citet{SGLD} showed that this distribution can be approximated using Stochastic Gradient Langevin Dynamics (SGLD) by injecting Gaussian noise to the parameter updates of SGD so that they do not collapse to just the MAP solution. This leads to the following parameter update:
\begin{equation}
\label{sgld}
\Delta \theta = -\frac{\ell}{2}\left(\frac{n}{N}\sum_{i=1}^{N}\nabla_{\theta} \log p(x_{i}|\theta)+ \nabla_{\theta}\log p(\theta)\right)+\sqrt{\ell}\epsilon,
 \;\epsilon \sim \mathcal{N}(0, I)  
\end{equation}
Note that when $\mathcal{D}$ is too large, it is to expensive to evaluate the log posterior $U(\theta):= \log p(\mathcal{D}|\theta)+\log p(\theta)$, for all the data points at each iteration. Hence, SG-MCMC methods use a mini-batch gradient to approximate $\nabla_{\theta}U(\theta)$ with an unbiased estimate  $\nabla_{\theta}U(\theta)\approx n\nabla_{\theta}\tilde{U}(\theta)$, where $\nabla_{\theta}\tilde{U}(\theta):=\frac{1}{N}\sum_{i=1}^{N} \nabla_{\theta} \log p(x_{i}|\theta)+ \frac{1}{n}\nabla_{\theta}\log p(\theta)$. In particular, note that the log prior \textit{scales} with the \textit{dataset size} at each iteration.  

SGHMC \citep{SGHMC} is an improved counterpart of SGLD which introduces a momentum variable $\mathbf{m}$. The posterior sampling is done using the following update rule:
\begin{align}
\label{sghmc}
\mathbf{m} & = \beta \mathbf{m} -\frac{\ell}{2}n\nabla_{\theta}\tilde{U}(\theta)+ \sqrt{(1-\beta)\ell}\epsilon; \;\; \epsilon\sim \mathcal{N}(0,I)\\ \nonumber 
\theta & = \theta + \mathbf{m}
\end{align}
where $\beta$ is the momentum term. 
The convergence to the true posterior is ensured by Equations~\eqref{sgld} and \eqref{sghmc}, given that learning rate $\ell$ follows the Robbins-Munro conditions and decays towards zero \citep{SGLD}. \citet{cSGMCMC} showed that replacing the traditional decreasing learning rate schedule in SGHMC with a cyclical variant allows to explore multimodal posterior distributions and developed cSGHMC. In this paper we apply cSGHMC to take samples from the posterior distribution.

\section{Posterior over Representations}
To infer a posterior over the embeddings, we place a \textit{prior} $\alpha$ over the parameters $\theta$ of the online network $\mathcal{F}_{\theta}$. By placing a distribution over $\theta$, we induce a distribution over an infinite space of online networks $\mathcal{F}_{\theta}$. This results in a distribution over embeddings $Z_{\theta}$. Sampling from this distribution corresponds to sampling from the following conditional posterior:
\begin{equation}
\label{posterior}
 p(\theta|X) \propto p(X|\theta)p(\theta|\alpha), 
\end{equation}
where $X$ is a mini-batch. Equation \eqref{posterior} can be interpreted intuitively as follows. We sample weights $\theta$ from the prior $p(\theta|\alpha)$. By conditioning on this sample of weights, we construct a specific online network $\mathcal{F}_{\theta}$. This network is then utilized to generate an embedding $Z_{\theta}$ by minimizing the loss function $\mathcal{L}^{\text{BYOL}}_{\theta, \xi}$.
\paragraph{Implementation Details.}
In practice, when considering the online network $\mathcal{F}_{\theta}(X) = f_{\theta}\circ g_{\theta}(\phi_{\theta} (t(X)))$, applied on a mini-batch $X$ transformed using data augmentation $t \sim \mathcal{T}$, we place a prior $\alpha$ on the parameters $\theta$ of the encoder $\phi_{\theta}$. As a prior $\alpha$, we assume isotropic Guassian distribution $\mathcal{N} (0, I)$. Subsequently, we compute 
$p(X|\theta) = ||f_{\theta}(Y_{\theta})-Y_{\xi}||^{2}$ and 
$
\tilde{U}(\theta) = \log p(X|\theta) + \frac{1}{n} \log p(\theta|\alpha)
\label{eq:postmin}
.$
\footnote{For simplicity of notations we ignore normalization and symmetrization}
Note that \cite{cont_ICA} indicated that contrastive learning inverts the data generating process and investigated the connection between contrastive learning and identifiability in the form of nonlinear Independent Components Analysis (ICA). Specifically, they showed that contrastive loss can be interpreted as the cross entropy between the (conditional) ground-truth and inferred latent distribution \citep{cont_ICA}
\footnote{For a more detailed and precise discussion on this matter, we refer to \cite{cont_ICA}}. This implies that considered loss can be interpreted as a negative log-likelihood, and by utilizing cSGHMC, we can efficiently obtain samples from the posterior using the update rule described in Equation \eqref{sghmc}. 
Our proposed method for sampling from the posterior distribution over embeddings $Z_{\theta}$ is outlined in Algorithm \ref{alg:Bself}. 
The algorithm generates samples from the posterior over the parameters $\theta$ of the online network $\mathcal{F}_{\theta}$. This yields a distribution over embeddings $Z_{\theta}$, as we compute the gradients of the loss with respect to the sampled parameters $\theta$.      

\citet{coldpos} indicated that tempering helps improve performance for Bayesian inference where $p(\theta|\mathcal{D}) \propto \exp (-U(\theta)/T)$, and $T < 1$ is a temperature. Here we also use cold posterior and choose $T$ by tuning on validation set. More details can be found in Appendix A.  

Our proposed probabilistic approach represents a natural Bayesian extension of MAP optimization, encompassing the advantages of uncertainty estimation. In fact, by performing MAP optimization using SGD in Algorithm \ref{alg:Bself} instead of posterior sampling, one approximates the entire \textit{posterior} distribution over $\theta$ with a single \textit{point estimate}, thus disregarding the richness of the full posterior. 
\begin{algorithm}
\caption{Probabilistic Self-Supervised Learning }
\label{alg:Bself}
\begin{algorithmic}
\Require $\ell $ initial learning rate, $\beta$ momentum term, $K$ number of training iterations  
\Ensure sequence $\theta^{1}, \theta^{2}, ...$
\For{$k=1:K$}
    \State sample mini-batch $X=\{x_{i}\}_{i=1}^{N}$ and augmentations $t\sim \mathcal{T}, t^{\prime}\sim \mathcal{T} $

    \State compute $\tilde{U}(\theta)$ for mini-batch $X$ 
    \State $\ell \gets C(k)\ell$ \Comment{update learning rate using cyclic modulation}
    \State $\mathbf{m} \gets \beta \mathbf{m}-\frac{\ell}{2} n\nabla_{\theta}\tilde{U}(\theta)+ \sqrt{T(1-\beta)\ell}\epsilon $
    \hspace{0.5cm} $\epsilon\sim \mathcal{N}(0,I)$
    \State $\theta \gets \theta+\mathbf{m}$
    \Comment{update parameters $\theta$ using Equation \eqref{sghmc}}
    \State $\xi \gets \tau \xi +(1-\tau)\xi$
\If{end of cycle}
    \State yield $\theta$
\EndIf
\EndFor
\end{algorithmic}
\end{algorithm}
\paragraph{Marginalizing over representations:}
After completing the pre-training phase, we can proceed to marginalize the posterior distribution over $\theta$ for downstream tasks. To compute the predictive distribution for a new instance $x^{*}$ we use a model average over all collected samples with respect to the posterior over $\theta$.
\begin{equation}
\label{marginal}
p(y^{*}|x^{*},\mathcal{D}) = \int p(y^{*}|x^{*},\theta) p(\theta|\mathcal{D}) d\theta 
\approx \frac{1}{S} \sum_{s=1}^{S} p(y^{*}|x^{*},\theta^{(s)}), \theta^{(s)} \sim p(\theta|\mathcal{D}) 
\end{equation}
We will observe that this model average significantly enhances performance, calibration, and out-of-distribution detection in downstream tasks. Additionally, by possessing samples from the posterior, we can compute the entropy for a given instance $x^{*}$, thereby providing an estimation of uncertainty in the predictive space: 
\begin{equation}
\mathcal{H}(y^{*}|x^{*}, \mathcal{D}) = -\sum_{c\in C} p(y^{*}|x^{*}, \mathcal{D}) \log p(y^{*}|x^{*}, \mathcal{D}) \nonumber
\end{equation}

\begin{figure}[t]
\centering

\begin{subfigure}[b]{0.3\textwidth}
\centering
\includegraphics[width=\textwidth]{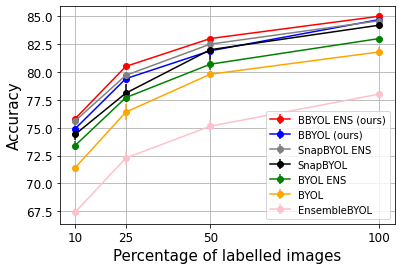}
\caption{CIFAR-10}
\label{in_vivo_10_dice}
\end{subfigure}
\hfill
\begin{subfigure}[b]{0.3\textwidth}
\centering
\includegraphics[width=\textwidth]{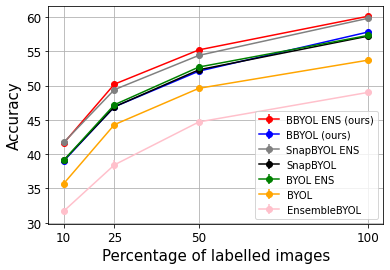}
\caption{CIFAR-100}
\label{in_vivo_2_dice}
\end{subfigure}
\hfill
\begin{subfigure}[b]{0.3\textwidth}
\centering
\includegraphics[width=\textwidth]{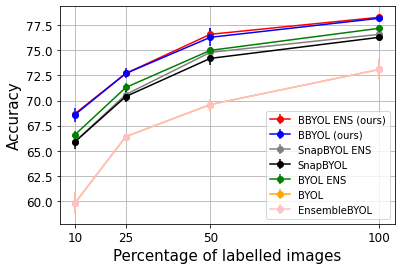}
\caption{ImageNet-10}
\label{dice_D}
\end{subfigure}
\\
\centering
\begin{subfigure}[b]{0.3\textwidth}
\centering
\includegraphics[width=\textwidth]{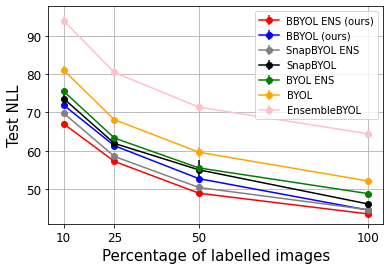}
\caption{CIFAR-10}
\label{in_vivo_10_dice}
\end{subfigure}
\hfill
\begin{subfigure}[b]{0.3\textwidth}
\centering
\includegraphics[width=\textwidth]{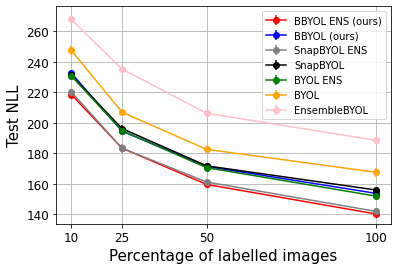}
\caption{CIFAR-100}
\label{in_vivo_2_dice}
\end{subfigure}
\hfill
\begin{subfigure}[b]{0.3\textwidth}
\centering
\includegraphics[width=\textwidth]{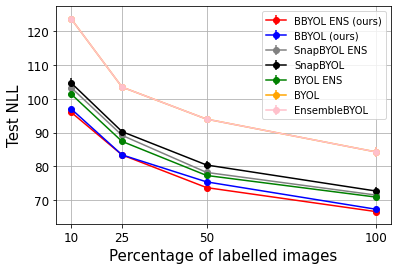}
\caption{ImageNet-10}
\label{dice_D}
\end{subfigure}
\caption{Performance comparison. First row indicates improvement in accuracy. The second row indicates improvement in calibration. Bayesian approaches outperform MAP estimation in terms of both accuracy and calibration. We use pre-trained model on STL-10 for CIFAR-10 and CIFAR-100. For ImageNet-10, a pre-trained model on Tiny-ImageNet is used. }
\label{acc_nll_semi}
\end{figure}

\section{Experiments}
\label{sec:exp}

In this section, we present our experimental results. We evaluate the performance and efficiency of the proposed method on several tasks including semi-supervised learning and out-of-distribution detection. First, we describe our experimental setup. Then, we evaluate our model using semi-supervised setting, and lastly, we evaluate our model using out-of-distribution examples. We implemented the code in PyTorch \citep{pytorch} and the link to our code is available in the supplementary material.  

\subsection{Experimental Setup}
\paragraph{\bf{Datasets}}
For pre-training phase, we pre-train all models on two image datasets STL-10 \citep{STL-10} and Tiny-ImageNet \citep{Tiny-ImageNet}. For STL-10, its 100,000 unlabeled samples are used for pre-training. For downstream task we conduct our experiments on four image classification datasets: CIFAR-10, CIFAR-100 \citep{cifar10}, STL-10 and ImageNet-10 \citep{ImageNet-10}. A brief description of these datasets is summarized in Table \ref{table:dataset}. For all datasets pre-trained models are fine-tuned on Train set and evaluated on Test set, except for ImageNet-10 that the Validation set  is used for evaluation, since the Test set of this dataset dose not have ground-truth labels. 
\begin{table}[b] 
\caption{A summary of datasets used for evaluations.} 
\centering
\label{table:dataset}
\begin{tabular}{c c c c c}
\toprule
Dataset & Split & Samples & Classes \\
\midrule
CIFAR-10  & Train+Test & 60000 & 10   \\
CIFAR-100  & Train+Test & 60000 & 100 \\
STL-10 & Train+Test & 13000  & 10  \\
ImageNet-10 & Train+Validation & 13000  &  10  \\
\bottomrule
\end{tabular}
\end{table}

\paragraph{Implementation Details}
We adopt ResNet-18 \citep{resnet18} as an encoder for the self-supervised learning model. Following the original setting of BYOL, we use 2-layer MLPs as the projection and prediction heads. We apply the standard ResNet without modification on the input images of original sizes for all datasets given in Table \ref{table:dataset} which produces a feature vector of size $512$ for each sample. We refer this feature vector as \textit{representation} or \textit{embedding}. We use the same set of data augmentations in \cite{BYOL} on both datasets for pre-training, consists of random cropping and resizing with a random horizontal flip, followed by a color distortion and a grayscale conversion.
\begin{figure}[!t]
     \centering
     \begin{subfigure}[b]{0.45\textwidth}
         \centering
         \includegraphics[width=0.9\textwidth]{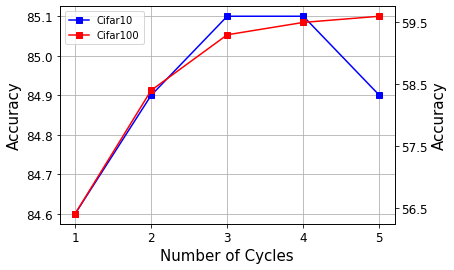}
         \label{acc_cyc}
     \end{subfigure}
     \hfill
     \begin{subfigure}[b]{0.45\textwidth}
         \centering
         \includegraphics[width=0.9\textwidth]{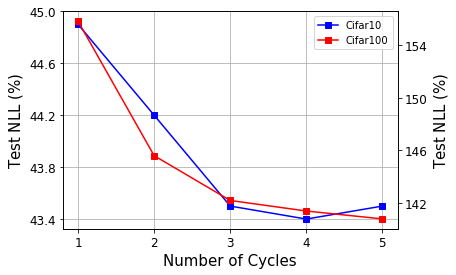}
         \label{nll_cyc}
     \end{subfigure}
        \caption{The effect of ensemble size on CIFAR-10 and CIFAR-100 in BBYOL ENS: (a) Testing accuracy ($\%$) (b) Testing NLL ($\%$) as a function of the number of cycles. Adding more embeddings improves accuracy and calibration.}
        \label{ens_size}
\end{figure}

\paragraph{Evaluation Metrics}
Two widely-used metrics including Accuracy (ACC), and Negative Log Likelihood (NLL) are utilized to evaluate our method. Higher value of ACC indicates better performance of the model and lower value of NLL indicates better calibration.

\paragraph{Baselines}
In order to demonstrate the effectiveness of our proposed probabilistic approach we conduct a comparative analysis with several methods including: (i) BYOL: MAP estimation trained with SGD; (ii) BYOL ENS: stochastic optimization ensemble method; (iii) SnapBYOL: MAP estimation trained with SGD and cyclical stepsize schedule; (iv) SnapBYOL ENS: a stochastic optimization ensemble method with a cyclical stepsize schedule and  (v) EnsembleBYOL: an ensemble of BYOL trained with SGD from scratch for different random initialization. In the aforementioned methods, when we utilize only the last embedding in downstream task we refer to the model as BYOL, SnapBYOL and BBYOL. However, when we perform marginalization over embeddings, we adopt BYOL ENS, SnapBYOL ENS and BBYOL ENS. It is worth nothing that EnsembleBYOL also signifies marginalizing over embeddings. 

For training BYOL we employed SGD optimizer with a fixed learning rate schedule. For SnapBYOL we utilized SGD optimizer with a cyclic stepsize schedule. In training BBYOL, we followed the training procedure outlined in Algorithm \ref{alg:Bself}, where we use a cyclic stepsize schedule \citep{cSGMCMC} with cycle length of $ 50$. All models are trained from scratch for $1000$ epochs. In BBYOL and SnapBYOL, we collect 1 sample at the end of each cycle for the last 4 cycles resulting in a total of 4 samples. In BYOL, we take 4 samples on last 200 epochs, maintaining a regular interval of 50 epochs between each sample. To ensure consistency in the training budget across all methods, we trained EnsembleBYOL by employing the SGD optimizer with a fixed learning rate for 250 epochs, using four different random seeds. The curve in the plot indicates marginalizing over 4 embeddings. Other training and baseline hyperparameters are provided in Appendix A.  


The experiments are carried out on Nvidia A40 48 GB and it takes about 21 gpu-hours on STL-10, and 24 gpu-hours on Tiny-ImageNet. We repeat experiments for 3 random seeds and report average NLL and ACC over 3 runs with the standard error from the mean predictor. 

\subsection{Image Classification}
In this section we present the evaluation results of proposed method on a semi-supervised image classification task. In this task, the quality of learned representations are assessed by fine-tuning a pre-trained model on subsets of original training datasets with labels. We evaluate over a variety of downstream training set sizes and analyze the obtained gains in performance and calibration. We follow the semi-supervised protocol in \citep{BYOL} and provide a detailed description of hyperparameters in Appendix A.

In Figure~\ref{acc_nll_semi}, we compare the above described methods across various dataset sizes in terms of accuracy and calibration. We observe the followings: (i) BBYOL consistently outperforms BYOL and SnapBYOL in all datasets in both metrics (the improvement of BBYOL over SnapBYOL in CIFAR-100 is relatively modest). (ii) Marginalizing over representations in BBYOL ENS improves performance and calibration compared to BBYOL. Marginalizing is more effective when the downstream task is more difficult for example in CIFAR-100. (iii) Marginalizing over representations in BYOL ENS and SnapBYOL ENS also improves performance. It is due to the nature of contrastive loss which induces diversity in the parameter space. Whenever the loss is not too high, marginalizing over these representations contributes to enhanced performance. However, even with this improvement, BBYOL ENS still achieves sizable gains in both performance and calibration over the baselines. (iv) Marginalizing is relatively more valuable on intermediate dataset sizes.      

Among above observations, Point (i) is particularly interesting, even if we do not want to use model averaging over representations due to a higher test-time cost, the last representation in a BBYOL trained using a Bayesian approach has significant better performance in accuracy and calibration compared to a MAP estimation. In Appendix B we provide additional evaluations with models pre-trained on Tiny-ImageNet.

 \begin{figure}[!t]
     \centering
     \begin{subfigure}[b]{0.45\textwidth}
         \centering
         \includegraphics[width=\textwidth]{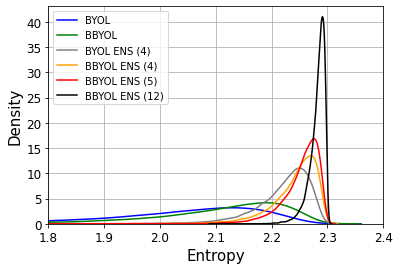}
         \label{acc_cyc}
     \end{subfigure}
     \hfill
     \begin{subfigure}[b]{0.45\textwidth}
         \centering
         \includegraphics[width=\textwidth]{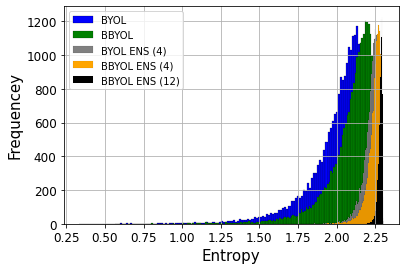}
         \label{nll_cyc}
     \end{subfigure}
     \caption{Histogram of the
predictive entropy on test examples from unknown data (SVHN), as we vary the ensemble size.}
\label{ood}
\end{figure}

\paragraph{Ensemble Size} 
In some applications, it may be beneficial to vary the
size of the ensemble dynamically at test time depending on available resources. Figure~\ref{ens_size} displays the performance of BBYOL ENS on CIFAR-10 and CIFAR-100 datasets as the effective ensemble size, is varied. Although ensembling more models generally gives better performance, we observe significant gains in accuracy and drops in NLL  when the second and third models are added to the ensemble. In most cases, an ensemble of two models outperforms the baseline model. In CIFAR-10, we have noticed a decline in performance upon introducing the fifth representation. This decrease can be attributed to the relatively high loss of this particular model, indicating that it may not be suitable for ensembling.
Therefore, we utilize the representations from the last four cycles in CIFAR-10 and CIFAR-100 for ensembling.

\begin{table*}
\centering
\caption{Results for OOD detection for various settings. Numbers in parentheses indicate number of embeddings. BBYOL ENS constantly outperforms all other methods for various number of ensembles. Underlined results indicate the results of our method and best results are in \textbf{bold face}.} 
\label{table:OOD}

\begin{tabular}{c c l l l}
\hline
In-Distribution & Out-of-Distribution &  Method & NLL $\downarrow$ & AUROC ($\%$) $\uparrow$  \\
\hline
&  & BYOL & $2.61\pm {\color{gray}\scriptsize 0.0}$ & $84.1\pm {\color{gray}\scriptsize 1.7}$ \\
&  & BBYOL & $\underline{2.53}\pm {\color{gray}\scriptsize 0.0}$ & $\underline{89.6}\pm {\color{gray}\scriptsize 0.7}$  \\
&  & BYOL ENS (4) & $2.38\pm {\color{gray}\scriptsize 0.0}$ & $93.6\pm {\color{gray}\scriptsize 0.4}$  \\
CIFAR-100 & SVHN  & BBYOL ENS (4) & $\underline{2.36}\pm {\color{gray}\scriptsize 0.0}$ & $\underline{95.7}\pm {\color{gray}\scriptsize 0.1}$  \\
 & & BYOL ENS (5) & $2.37\pm {\color{gray}\scriptsize 0.0}$ & $94.8\pm {\color{gray}\scriptsize 0.3}$  \\
 &  & BBYOL ENS (5) & $\underline{2.35}\pm {\color{gray}\scriptsize 0.0}$ & $\underline{96.4}\pm {\color{gray}\scriptsize 0.1}$ \\
 &  & BBYOL ENS (12) & $\mathbf{2.32}$ & $\mathbf{98.3}$ \\
\hline
&  & BYOL & $2.62\pm {\color{gray}\scriptsize 0.02}$ & $84.3 \pm {\color{gray}\scriptsize 1.4}$ \\
&  & BBYOL & $\underline{2.53}\pm {\color{gray}\scriptsize 0.0}$ & $\underline{89.4} \pm {\color{gray}\scriptsize 0.5}$ \\
&  & BYOL ENS (4)& $2.38\pm {\color{gray}\scriptsize 0.0}$ & $93.6\pm {\color{gray}\scriptsize 0.3}$ \\
CIFAR-100 & CIFAR-10 & BBYOL ENS (4) & $\underline{2.37}\pm {\color{gray}\scriptsize 0.0}$ & $\underline{95.8} \pm {\color{gray}\scriptsize 0.2}$  \\
 &   & BYOL ENS (5) & $2.37 \pm {\color{gray}\scriptsize 0.01}$ & $94.8\pm {\color{gray}\scriptsize 0.3}$ \\
 &  & BBYOL ENS (5) & $\underline{2.35}\pm {\color{gray}\scriptsize 0.0}$ & $\underline{96.4} \pm {\color{gray}\scriptsize 0.1}$ \\
 &  &  BBYOL ENS ($12$) & $\mathbf{2.31}$ & $\mathbf{98.2}$  \\
\hline
\end{tabular}
\end{table*}

\subsection{Out-of-Distribution Detection}
To further analyze the effectiveness of the proposed probabilistic approach compared to the MAP estimation, we consider the out-of-distribution (OOD) detection task \citep{cSGMCMC}. 
In this task, a model trained on known data is evaluated on unseen data. For the unseen data, we expect the model indicates low probability and max entropy \citep{cSGMCMC}. It implies that the mode of the predictive entropy's histogram focuses at higher value. Moreover, we assess the quality of the predictive uncertainty using two quantitative metrics, NLL and the area under the receiver operating characteristic curve (AUROC) \citep{AUROC}, a higher value of AUROC indicates a better detector.   

We consider two datasets CIFAR-10 and SVHN \citep{SVHN} as OOD datasets. A pre-trained model on STL-10, is fine-tuned on CIFAR-100 and evaluated on SVHN and CIFAR-10. Figure~\ref{ood} presents the histogram of the predictive entropy for SVHN (out-of-distribution). The histogram of the  predictive entropy for CIFAR-10 had the same distribution, so we just included SVHN. 

We see that the uncertainty estimates from BBYOL and BBYOL ENS are better than the other methods, as the mode of histogram focuses at higher values. BYOL ENS also improves uncertainty estimate on unseen data compared to BYOL but still achieves less entropy than BBYOL ENS. Moreover the predictive uncertainty improves on unseen data, as the ensemble size increases reaching to the highest value in BBYOL ENS (12), where we take 12 samples from last 4 cycles (3 samples per cycle). It indicates that embeddings produced by sampling from the posterior in BBYOL come from different modes and provide different characterization of training data. When testing on unseen data, each mode provides different predictions on unseen data leading to max disagreement and higher entropy. 

 We also report the quantitative results for NLL and AUROC in Table~\ref{table:OOD} as the ensemble size is varied. BBYOL ENS (5) indicates marginalizing over 5 embeddings collected from last 5 cycles. Consistent with our previous results BBYOL improves BYOL in terms of both calibration (lower NLL) and AUROC (higher) by large margin ($5.5 \%$ in AUROC only for last embedding and $14.2\%$ when we marginalize over 12 embeddings). The improvement in calibration and AUROC consistently increases by increasing the number of ensemble size. 
 
\section{Conclusion}
\label{others}
In this paper, we propose a novel approach that challenges the traditional Maximum A Posteriori (MAP) solution for learning representations, advocating instead for the utilization of Bayesian methods. Our primary objective is to thoroughly explore the posterior distribution over the representations and investigate the potential advantages offered by incorporating probabilistic sampling techniques into representation learning. By deviating from the traditional MAP approach, we aim to shed light on the extensive benefits and valuable insights that can be gained from embracing Bayesian approaches in representation learning.
To achieve this, we employ a powerful SG-MCMC method designed to capture the multi-modal posterior distribution. Through extensive experiments, we have obtained compelling findings that underscore the distinctiveness of samples derived from the posterior distribution. This distinctiveness translates into remarkable improvements across multiple metrics, including accuracy, calibration, and uncertainty estimation, in downstream tasks. By embracing the richness of the posterior, we empower models to better capture the inherent complexity and nuances of the underlying data.

\section{Broader Impact}
Incorporating Bayesian approaches into self-supervised learning has the potential to significantly impact the field across multiple dimensions. One prominent advantage is the ability to introduce the concept of uncertainty, not only in the predictive space but also in the embedding space. Exploring the relationship between these two forms of uncertainty presents an intriguing avenue for analysis, offering valuable insights into their interplay and potential implications. Understanding how uncertainty manifests in both prediction and embedding spaces can contribute to a more comprehensive understanding of the underlying data and enhance the robustness and reliability of self-supervised learning algorithms.


\bibliography{ref}
\end{document}


\bibliographystyle{abbrvnat}

\maketitle
In this supplementary material, we provide (1) further details in implementation of proposed method and baselines in pre-training as well as downstream phase. (2) We also provide additional experimental results in semi-supervised image classification task \footnote{Our code is available under the follwoing link: \url{https://drive.google.com/file/d/1TVfe1ZGache2qS3Z_-k8ypMYb9zGoUSH/view?usp=share_link}}.

\appendix
\label{parameter description}
\section{Implementation Details}
\label{parameter_desc}
\paragraph{Pre-training} To select hyperparameters, we hold out a subset of training set for validation. After selecting the optimal hyperparameters from the validation set, we retrain the model using the selected parameters on both the training and validation images.

\paragraph{BBYOL} For training BBYOL we use cSGHMC~\citep{cSGMCMC} with cyclic learning rate schedule of length $50$ epochs. In accordance with \citep{cSGMCMC}, we adopt normal prior $\mathcal{N}(0,I)$ for the parameters of the online network. As stated in the main paper, scaling the prior with the dataset size is necessary for training cSGHMC, given that we evaluate it on a minibatch of data. For injecting Gaussian noise, we swapped over epochs $\{35,40,45\}$ and select epoch $40$. For the initial learning rate we swapped over $\{0.1,0.2,0.3\}$ and set the initial learning rate to $0.2$. The batch size is set to $256$ and we use momentum term of $0.9$. As described in \citep{cSGMCMC}, tempering helps improve performance for Bayesian inference with neural networks. So, we swapped over $\{0.1,0.01\}$ and set temperature to $0.1$. 
  
\paragraph{BYOL} For training BYOL we use SGD optimizer with a fixed learning rate schedule and momentum term $0.9$. For the initial learning rate we swapped over $\{0.0003,0.003\}$ and set the learning rate to $0.003$. We found that using weight decay destroys representations so no weight decay is used. Other parameters are as the same as BBYOL. 

\paragraph{SnapBYOL} To train SanpBYOL, we employ the SGD optimizer with a cyclic learning rate schedule of length $50$ epochs. 
For the initial learning rate, for STL-10 we swapped over $\{0.1, 0.09, 0.07, 0.03, 0.02, 0.01, 0.009\}$ and set the initial learning rate to $0.01$. For Tiny-ImageNet we swapped over $\{0.03, 0.02, 0.01, 0.009, 0.008, 0.007, 0.006\}$ and set the initial learning rate to $0.006$. It is worth noting that for learning rates higher than this, the model does not converge when applied to Tiny-ImageNet.
Similar to BYOL, we did not employ weight decay, and all other parameters remain consistent with those of BYOL.

\paragraph{EnsembleBYOL} To train EnsembleBYOL, we utilized the same parameters as BYOL while incorporating a normal prior of $\mathcal{N}(0, I)$ on the parameters of the online network. We trained the model for four different random seeds, each for 250 epochs. 

\paragraph{Semi-supervised image classification} 
For fine-tunning, we follow the protocol of \citet{Byol}. We first initialize the network with the parameters of the pre-trained representation, and fine-tune it with a subset of original datasets with labels. We do not use any data augmentation during fine-tunning.  We optimize the cross-entropy loss using SGD with Nesterov momentum with batch size of $80$, and a momentum of $0.9$. We sweep over the learning rate $\{2e-5,1e-5,1e-4,2e-4,3e-4,4e-4,5e-4\}$, weight decay $\{0, 5e-4\}$ and the number of epochs $\{50, 60\}$ and select the hyperparameters achieving the best performance on our local validation set to report test performance. Table \ref{tab:fp} describes parameters for each dataset.

\begin{table}[h]
\centering
\caption{Parameters used in fine-tunning.} 
\label{tab:fp}
$\\$
\begin{tabular}{l l c c c}
\hline
 Pre-training & Fine-tuning & Split~($\%$) & learning rate & weight decay  \\
\hline
 &  & $100$ &$2e-4$ & $0$ \\
 & CIFAR-10 & $50$ &$5e-4$ & $0$ \\
&  & $25$ &$5e-4$ & $0$ \\
 &  & $10$ &$5e-4$ & $0$ \\
\hline
&  & $100$& $5e-4$ & $0$  \\
STL-10 & CIFAR-100 & $50$& $3e-4$ & $0$  \\
&  & $25$& $4e-4$ & $0$  \\
& & $10$& $5e-4$ & $0$  \\
\hline
&  & $100$& $5e-4$ & $0$  \\
& STL-10 & $50$& $5e-4$ & $0$  \\
&  & $25$& $5e-4$ & $0$  \\
& & $10$& $5e-4$ & $0$  \\
\hline
\hline
&  & $100$& $2e-4$ & $0$  \\
 & CIFAR-10 & $50$& $5e-4$ & $0$  \\
&  & $25$& $4e-4$ & $0$  \\
& & $10$& $4e-4$ & $0$  \\
\hline
&  & $100$& $5e-4$ & $0$  \\
 Tiny-ImageNet & CIFAR-100 & $50$& $5e-4$ & $0$  \\
&  & $25$& $5e-4$ & $0$  \\
& & $10$& $5e-4$ & $0$  \\
\hline
&  & $100$& $2e-4$ & $0$  \\
 & ImageNet-10 & $50$& $5e-4$ & $0$  \\
&  & $25$& $3e-4$ & $0$  \\
& & $10$& $5e-4$ & $0$  \\
\hline
\end{tabular}
\end{table}

\section{Additional Experimental Results}
\begin{figure}[!h]
\centering
\begin{subfigure}{0.4\textwidth}
\includegraphics[width=\textwidth]{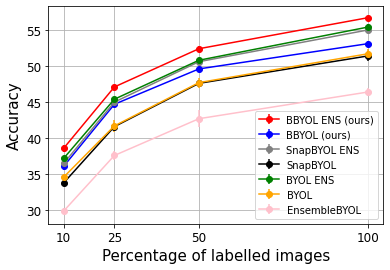}
\caption{CIFAR-100}
\label{fig:sub1}
\end{subfigure}\hskip1ex
\begin{subfigure}{0.4\textwidth}
\includegraphics[width=\textwidth]{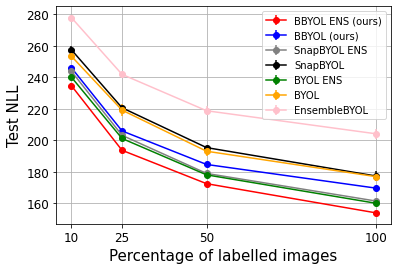}
\caption{CIFAR-100}
\label{fig:sub2}
\end{subfigure}
\caption{Performance comparison. Bayesian approach outperforms MAP estimation in terms of both accuracy and calibration.}
\label{extra_plot}
\end{figure}

In this section we provide results obtained from pre-training on Tiny-ImageNet and fine-tuned on CIFAR-10 and CIFAR-100. We pre-train all models with parameters described in Section~\ref{parameter_desc} and take $4$ samples at last $200$ epochs for BYOL, BBYOL and SnapBYOL. For EnsembleBYOL, we marginalise over representations obtained from four pretrained models, each trained with different random seeds. The results in semi-supervised image classification on CIFAR-10 and CIFAR-100 are illustrated in Figure~\ref{extra_plot}. Consistent with our results from pre-training on STL-10, BBYOL and BBYOL ENS outperform their MAP estimation counterparts in both metrics and various dataset sizes.
\newpage
\begin{figure}[!h]
\centering
\begin{subfigure}{0.4\textwidth}
\includegraphics[width=\textwidth]{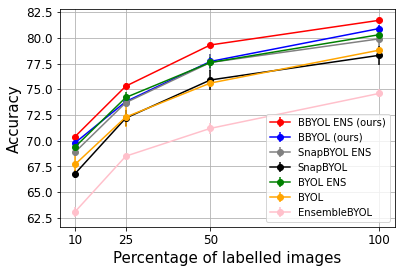}
\caption{CIFAR-10}
\label{fig:sub1}
\end{subfigure}\hskip1ex
\begin{subfigure}{0.4\textwidth}
\includegraphics[width=\textwidth]{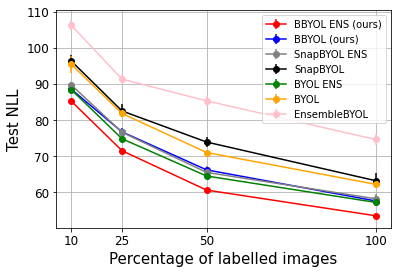}
\caption{CIFAR-10}
\label{fig:sub2}
\end{subfigure}
\end{figure}

\bibliography{ref}